\documentclass[conference]{IEEEtran}
\IEEEoverridecommandlockouts

\usepackage{graphicx}
\usepackage{xcolor}
\usepackage{color}
\usepackage{url}
\usepackage{times}
\usepackage{xspace}
\usepackage{algorithm}
\usepackage{algorithmic}
\usepackage{amssymb}
\usepackage{amsmath}
\usepackage{amsfonts}
\usepackage{graphicx}
\usepackage{multicol}
\usepackage{balance}
\usepackage{enumitem}
\usepackage{framed}
\usepackage{array,tabularx}
\usepackage{booktabs,makecell}
\usepackage{anyfontsize}

\usepackage{balance}  

\newcommand{\hide}[1]{}

\newcommand{\vy}{\mathbf{Y}\xspace}

\newcommand{\cumfeature}{\texttt{CumF}\xspace}
\newcommand{\emb}{\texttt{Emb}\xspace}
\newcommand{\lp}{\texttt{LabelProp}\xspace}
\newcommand{\cumlp}{\texttt{CumF+LP}\xspace}
\newcommand{\all}{\texttt{All}\xspace}
\newcommand{\emblp}{\texttt{Emb+LP}\xspace}

\newcommand{\mlp}{\textsc{MLP}\xspace}
\newcommand{\cnn}{\textsc{CNN}\xspace}

\newcommand{\lr}{\textsc{LogisticRegression}\xspace}
\newcommand{\gbdt}{\textsc{GBDT}\xspace}
\newcommand{\xgboost}{\textsc{XGBoost}\xspace}
\newcommand{\linearsvm}{\textsc{LinearSVM}\xspace}
\newcommand{\nn}{\textsc{MLP}\xspace}
\newcommand{\resnet}{\textsc{ResNet}\xspace}

\makeatletter
\def\@copyrightspace{\relax}
\makeatother

\begin{document}


\title{Large-scale Gender/Age Prediction of Tumblr Users}


\author{
%
%


\IEEEauthorblockN{Yao Zhang\IEEEauthorrefmark{1}\thanks{\IEEEauthorrefmark{1}The work was done when the author was at Yahoo Research.},
Changwei Hu\IEEEauthorrefmark{2}, 
Yifan Hu\IEEEauthorrefmark{2},
Tejaswi Kasturi\IEEEauthorrefmark{2},\\
Shanmugam Ramasamy\IEEEauthorrefmark{3},
Matt Gillingham\IEEEauthorrefmark{3},
 and Keith Yamamoto\IEEEauthorrefmark{3}}
\IEEEauthorblockA{\IEEEauthorrefmark{1}LinkedIn
\IEEEauthorrefmark{2}Yahoo Research
\IEEEauthorrefmark{3}Verizon Media\\
Email: \IEEEauthorrefmark{1}{yaozhang@linkedin.com}
\IEEEauthorrefmark{2}\{{changweih, yifanhu, kasturit}\}{@verizonmedia.com}}
\IEEEauthorrefmark{3}\{{sramasamy8, mgilling,  yamamoto}\}{@verizonmedia.com}}



\maketitle

\begin{abstract}
Tumblr, as a leading content provider and social media, attracts 371 million monthly visits, 280 million blogs and 53.3 million daily posts\footnote{ \url{http://expandedramblings.com/index.php/tumblr-user-stats-fact/}}.The popularity of Tumblr provides great opportunities for advertisers to promote their products through sponsored posts. However, it is a challenging task to target specific demographic groups for ads, since Tumblr does not require user information like gender and ages during their registration. Hence, to promote ad targeting, it is essential to predict user's demography using rich content such as posts, images and social connections. In this paper, we propose graph based and deep learning models for age and gender predictions, which take into account user activities and content features.  For graph based models, we come up with two approaches, network embedding and label propagation, to generate connection features as well as directly infer user's demography. For deep learning models,  we leverage convolutional neural network (CNN) and multilayer perceptron (MLP) to prediction users' age and gender. Experimental results on real Tumblr daily dataset, with hundreds of millions of active users and billions of following relations, demonstrate that our approaches 
significantly outperform the baseline model, by improving the  accuracy relatively by $81\%$ for age, and the AUC and accuracy by $5\%$ for gender.
\end{abstract}

\section{Introduction}

Online social media has become a ubiquitous part of our daily life, which allows us to easily share ideas/contents with other users, discuss social events/activities, and get connected with friends. Tumblr, with over 280 million blogs\footnote{A Tumblr user typically corresponds to one primary blog, so in this paper users and blogs are used interchangeably.} and 130 billion blog posts, is one of the most popular social media apps. The rich content including text, images, and videos, provide great opportunities for advertisers to champion their products to specific groups. In particular, Tumblr offers ``native advertisement" that allows advertisers to present their sponsored posts on the users'' interface.   Native advertising has gained over 3 billion paid ad impressions in 2015 since it was started in 2012~\cite{Grbovic:2015}.  However, unlike other social networks,  Tumblr does not ask for some of the user demographic information, such as gender, during registration. Even though age is a required input during registration, it is difficult to
verify the accuracy of this self-declared information.
This makes it a challenge to precisely target ads for a group of users with age and gender related demography profiles.
 To improve the performance of user specific ad targeting, it is important for Tumblr to infer users' age and gender from rich content generated by users. 

 Several attempts have  been made for this task, e.g., Grbovic et al.~\cite{Grbovic:2015} proposed a gender and interest targeting framework that leverages user-generated data. Their model has lifted user engagement of ads, however, they did not take into consideration age prediction, which is one of the key factors for targeted ads. Furthermore, their gender prediction models only use features from blog contents and user activities (the so-called ``cumulative features"). We believe that in addition to the cumulative features, user interactions, represented by the Tumblr following graph, can also be used for age and gender prediction, as users who followed each other tend to have similar interests or background. It is worth mentioning that the model in~\cite{Grbovic:2015} does use blogs a user follows as a categorical feature for a linear classifier. However, it only exploits 1-hop neighbor information of the Tumblr following graph. The rich network structure provides further indications on how users interact. Furthermore, with the development in deep learning, advanced models like convolutional neural network (CNN) and multilayer perceptron (MLP) can typically produce better performance, which the past work like~\cite{Grbovic:2015} does not explore.

Hence, to achieve better performance for age and gender prediction, in this paper, we propose a graph based approach with deep learning techniques that 
leverages the multitude of information encoded by the Tumblr following graph.

The main contributions of our paper include:

\begin{itemize}
	\item \emph{Data:}  We leverage rich cumulative features including user activities and post contents. Furthermore, we construct a large Tumblr following graph with hundreds of millions of vertices and billions of edges.

	\item \emph{Graph base methods:} We apply graph embedding and label propagation techniques to (1) generate rich features from the Tumblr following graph through node embedding, which is shown to be useful to improve users demography prediction; (2) directly leverage the  label propagation algorithm to boost the prediction performance. 
	
	\item \emph{Deep learning:} We apply deep learning models including CNN and MLP to further improve the performance of age and gender prediction. 

	\item \emph{Evaluation:}  We conduct empirical studies to show that the graph based and deep learning approaches can improve the AUC and accuracy performance by $5\%$ relatively  for gender prediction and {the accuracy performance by $81\%$ for age prediction}, compared to the baseline model~\cite{Grbovic:2015}, and
	classifiers like \gbdt, \xgboost, etc. 
	\end{itemize}
The rest of this paper is organized as follows. Section~\ref{sec_data} describes the Tumblr data, and Section~\ref{sec:proposed} discusses our proposed methods, including network embedding, label propagation and deep learning. Empirical studies are shown in Section~\ref{section_results}.
Finally we summarize the related work in Section~\ref{sec:related}, and conclude our work with future directions in Section~\ref{sec:conclusion}.

\section{Preliminary}
\label{sec_data}

In this section, we discuss the rich Tumblr data we use, and the labels we create.

\subsection{Tumblr Data}

We use both following graph and cumulative features. 
\smallskip

\noindent \textbf{Following Graph.} In Tumblr, users can follow other users, which forms the Tumblr following graph. Formally, we define $G(V, E)$ as the following graph, where $V$ is the set of users and $E$ is the set of following relations. Note that even though the following graph is directed in Tumblr, we assume $G$ is an unweighted undirected graph, as our study focuses on age/gender prediction where we assume two users with a following relation tend to share mutual interests in their background. In this paper, $G$ consists of hundreds of millions of nodes and billions of edges.

\smallskip

\noindent \textbf{Cumulative Features.} We utilize ``cumulative features"~\cite{Grbovic:2015}, which contain blog content and activities including {\em music} (artist), {\em follow} (blogs followed), {\em likes} (of posts), {\em photo captions}, {\em post tags}, {\em blog titles}, and {\em blog descriptions}, etc. 

Note that the baseline
model~\cite{Grbovic:2015} only used cumulative features with \lr for the Tumblr age/gender prediction. It is interesting to study how the following graph can boost the performance. 

\subsection{Label Construction}

Tumblr does not ask for gender information when a user signs up. Furthermore, although it does ask for user age, no independent verification is done, and Tumblr believes that
the age information is unreliable. Therefore, it is a challenge to create ground truth labels. Grbovic et al.~\cite{Grbovic:2015} leveraged the US census data that associates people's names with gender to create gender labels. However, their approach suffers from a natural issue that some names can be neutral (e.g., ``Avery"). Instead, we use age and gender ground truth data provided internally at Yahoo,  which is at a large-scale and more reliable. This golden set provides us with good quality of ground truth label information. In total there are about $3$ million ground truth age and gender labels.

\section{Proposed Methods}
\label{sec:proposed}
In this section, we will discuss our approaches for age and gender prediction, including network embedding, label propagation and deep learning.
We first enrich the cumulative features with graph features generated by network embedding and label propagation, and then apply deep learning techniques for the prediction. In addition, we also use label propagation as a direct tool for the prediction, which is shown to be efficient in the experiments.

\subsection{Network Embedding}
\label{section_embedding}

When applying \lr on the cumulative features for the baseline
model, we found that {\em follow} and {\em like} are the two most important features, which suggests that blogs that a user follows/reads is a good indicator on the user's demographic. Intuitively, users tend to follow others' activities with similar age/gender. Using {\em follow} features directly is akin to using words as unigram features for natural language processing. It works to some extent, but fails to uncover underlying connections among blogs. For example, if two blogs have a large overlap of followers, though they are not directly followed by each other,  they are somewhat related. However, network embedding (mapping blogs to a high dimensional space based on their follow relations) can capture such relationships.

In this paper, we apply word2vec~\cite{Mikolov_2013_word2vec,pennington2014glove} to generate rich network features. Word2vec is one of the most popular embedding techniques to capture syntactic and semantic relationships of entities. It leverages a multilayer perceptron (MLP) with two architectures,  bag-of-words (BOW) and skip-gram. Levy and Goldberg~\cite{DBLP:conf/nips/LevyG14} found that the skip-gram model, which uses a word to predict its neighborhood words, is equivalent to a matrix factorization with the matrix to be factored containing the word-word point-wise mutual information. With this broad understanding of word2vec in mind, the technique is applicable  to tasks beyond those of traditional natural language processing tasks. It can be deployed to applications with entities and their co-occurrence patterns. Since blogs and their followers have such a pattern, we leverage word2vec for our task.

\smallskip
\noindent \textbf{Implementation details.} Network embedding has been well-studied recently. For example, DeepWalk~\cite{perozzi2014deepwalk} constructs ``sentences" of vertices by random walks, and then applies a word2vec model to sentences to get vector representation of vertices.  Vertices that appear frequently together in sentences tend to be similar. Our implementation is similar to LINE~\cite{Tang_2015_line}, where we create a sentence from a vertex and its neighbors. 
Specifically, each blog is combined with all the $k$ blogs it follows. This set of $k+1$ blogs are randomly permuted and is treated as one sentence. All together, the following graph $G$ produces hundreds of millions of ``sentences'' with billions of words, all of which are embedded into 50 dimensions using word2vec in the skip-gram mode. 

We use a 
word2vec implementation 
with a minimum word count of 5, which results in a dictionary size of 46 million. In order to use blog embedding successfully as features, we embed a majority of the blogs. For blogs without an embedding, we construct its embedding by averaging the embeddings of its neighbors that have an embedding. We denote the embedding algorithm as  \emb.

\subsection{Label Propagation}
\label{section_lp}

\emb is an efficient method to generate graph features. However, it is an unsupervised approach which does not take into account label information for training.  Next we develop semi-supervised learning algorithm based on label propagation (called \lp), which takes into account labels and  multi-hop neighborhood information. \lp serves as an approach to directly predict age/gender, as well as generate additional graph features. 
In addition, compared to the time consuming training of \emb for large graphs, \lp is a faster algorithm that can be naturally implemented in a distributed environment with a parallel fashion using Spark on Hadoop.

\lp spreads existing age/gender labels over the following graph. The intuition is that users tend to follow blogs with the similar background (e.g., ladies fashion). The age/gender of such a popular blog, defined by its followers, in turn defines the age/gender of its follower whose demographic information may be unknown to us. Next we will introduce the \lp algorithm and its variants, and then discuss how to leverage it for feature generation.

\smallskip
\noindent \textbf{The \lp Algorithm.} Given the Tumblr following graph $G(V, E)$, we denote the set of labeled users as $V_\ell=\{v_1, \dots, v_{\ell} \}$ where $V_\ell \subseteq V$, and the set of unlabeled users is denoted as $V_m$ where $V_m = V - V_\ell$. For $v_i \in V_\ell$, let $y_i$ be $v_i$'s label. The label propagation is an iterative process: at each iteration, a node's label propagates to its 1-hop neighbors. The process starts with $V_\ell$, and once a node gets labels from its neighboring nodes, it will propagate labels in the next iterations. The process ends when it converges or reach a pre-defined number of iterations. Note that during the process, we fix the labels of nodes in $V_\ell$, i.e., only nodes without labels at the beginning will iteratively update labels. 

Algorithm~\ref{alg:lp} shows the \emph{Label Propagation} (\lp) algorithm. We iteratively update label $y_i$ at each iteration (Line 4), where $N(i)$ is the set of neighboring nodes of $v_i$ that have labels at previous iterations. The algorithm ends until we reach $K$ iterations. Note that we also add a hyperparameter $\alpha \in [0, 1]$ to control contributions of neighboring labels. If $\alpha$ is low, we focus more on the neighboring information and vice versa.

\begin{algorithm}
	\caption{$\lp(G, V_\ell, \alpha, K)$}
	\label{alg:lp}
	\begin{algorithmic} [1]
		\REQUIRE $G$, $V_\ell$, $\alpha$, and iteration number $K$ 
		\STATE $k=1$
		\WHILE{$k \leq K$}
		\FOR {$v_i \in V_m$}
		\STATE ${y}_i^{(k+1)} := \alpha {y}_i^{(k)} + (1 - \alpha)\frac{1}{|N(i)|}  \sum_{j \in N(i)} y_j^{(k)} $
			\ENDFOR
			\STATE $k := k+1$
			\ENDWHILE
			\STATE return  $y_i^{(K)}$
		\end{algorithmic}
	\end{algorithm}

\hide{
At each iteration $k$, the label $y_i$ for node $v_i$ ($v_i \in V_m$) is updated as follows:

\begin{align}
\label{eqn:lp}
{y}_i^{(k)} = \alpha {y}_i^{(k-1)} + (1 - \alpha)  \sum_{j \in N(i)}  \frac{y_j^{(k-1)}} {| N(i) |} 
\end{align}
}

\noindent \textbf{Implementation details.} The label here can be either gender or age. For gender, we assume $y_i=1$ as female and $y_i=0$ as male. For age, we first split age into $7$ buckets $(<17, 18-24, 25-34, 35-44, 45-54, 55-64, 65+)$, and use one hot encoding to create a label vector with $7$ entries. We run \lp for each entry. It is natural to develop \lp in parallel, as each node updates its label by independently querying its neighbors at each iteration. We implement \lp under the Pregel framework~\cite{malewicz2010pregel} on Spark, which is a state-of-the-art message passing parallel model for large distributed graphs. 

Note that we do not run \lp until it converges, but stop at a predefined maximum number of iterations because of the scalability consideration (we may need a large amount of iterations to converge).  Furthermore, as we mentioned above, the motivation of \lp is that users that have following relations have similar demography. Hence, labels propagated to a node at a large number of iteration $k$, may ``contaminate" the prediction of $u$, as the propagated labels at $k$-th iteration can be different from the true label.

\hide{
\noindent \textbf{Variants of label propagation.}  We propose two variants of \lp in order to investigate how labels from nodes with different distance make impact on the label prediction.  The main idea is that instead of using a constant hyperparameter $\alpha$, we adaptively change the parameter. 
As the value of the iteration increases, we decrease the contribution of incoming labels from neighboring nodes. Hence, labels that are far away become less important in Algorithm~\ref{alg:lp} (Line 4). If they do not affect the performance, then the label prediction results with large iterations will remain competitive compared to the case with small iterations.
The followings are two alternative propagation strategies we propose:

\begin{enumerate}
\item $\beta$-strategy: ${y}_i^{(k+1)} := (1 -\beta^k) {y}_i^{(k)} + \beta^k  \sum_{j \in N(i)}  \frac{y_j^{(k)}} {| N(i) |} $.

\item $\gamma$-strategy: ${y}_i^{(k+1)} (\texttt{m/f}) :=  {y}_i^{(k)} (\texttt{m/f}) + \gamma \sum_{j \in N(i)}  \frac{y_j^{(k)}(\texttt{m/f})} {| N(i) |} $
\end{enumerate}

In the first strategy, $\beta \in [0, 1]$ and $k$ is the number of the iterations. As the iteration value increases, the impact of labels from the neighboring information decreases exponentially. In the second strategy, $\gamma \in [0, 1)$. We propagate male and female labels respectively for each node, and at the end normalize them as the final label for each node.  As shown in the experiments, these two strategies can empirically demonstrate that for a node $u$, labels from its close neighbors are more important than long-distance labels for gender prediction.
}

\smallskip

\noindent \textbf{Variants of label propagation.}  We propose two variants of \lp for the gender prediction in order to investigate how labels from nodes with different distance make impact on the label prediction. Note that it is straightforward to generalize the analysis to the multiclass age prediction.  The main idea is that instead of using a constant hyperparameter $\alpha$, we adaptively change the parameter. 
As the value of the iteration increases, we decrease the contribution of incoming labels from neighboring nodes. Hence, labels that are far away become less important in Algorithm~\ref{alg:lp} (Line 4). If they do not affect the performance, then the label prediction results with large iterations will remain competitive compared to the case with small iterations.
The followings are two alternative propagation strategies we propose:

\begin{enumerate}
\item $\beta$-strategy: ${y}_i^{(k+1)} := (1 -\beta^k) {y}_i^{(k)} + \beta^k  \sum_{j \in N(i)}  \frac{y_j^{(k)}} {| N(i) |} $.

\item $\gamma$-strategy: ${y}_i^{(k+1)} (\texttt{m/f}) :=  {y}_i^{(k)} (\texttt{m/f}) + \gamma \sum_{j \in N(i)}  \frac{y_j^{(k)}(\texttt{m/f})} {| N(i) |} $
\end{enumerate}

In the first strategy, $\beta \in [0, 1]$ and $k$ is the number of the iterations. As the iteration value increases, the impact of labels from the neighboring information decreases exponentially. In the second strategy, $\gamma \in [0, 1)$. We propagate male and female labels respectively for each node, and at the end normalize them as the final label for each node.  As shown in the experiments, these two strategies can empirically demonstrate that for a node $u$, labels from its close neighbors are more important than long-distance labels for gender prediction.

\smallskip

\noindent \textbf{Label Propagation as Features.} 
In \lp, we can directly learn labels by the iteration rule (Algorithm~\ref{alg:lp} Line 4). As shown in the next experiments, it provides high quality predictions compared to baselines. However, neither does it take into account the cumulative features, nor features from \emb. \emb generates graph features in an unsupervised manner without considering existing label information. To leverage the label information to improve the prediction performance, we use \lp to general features as well.

The idea of \lp as feature generators comes from the ensemble methods: we divide the labeled data uniformly at random into $N$ partitions. For partition $i$, we denote it as $V^i_\ell$. We run \lp on the whole graph with the labeled data $V^i_\ell$ for each $i$ respectively to get the learned label $y^i_u$, and concatenate learned labels. For each node $u$, label propagation features are represented as a vector $\vy_u = [y^1_u, \dots, y^N_u]^T$. For $u \in V_\ell$, if it is in partition $i$ (meaning we know its label $y_u$), instead of using $y_u$ as the feature at dimension $i$, we set $y^i_u = \frac{\sum_{j \neq i} y^j_u}{N-1}$. This is because directly putting $y_u$ as features will cause overfitting.

\subsection{Deep Learning Models}
\label{sec:deeplearning}

Deep learning models have been extensively studied due to their extraordinary performance in a variety of areas including computer vision, natural language processing, robotics, etc. The 
baseline model used the \lr method for age/gender prediction. In this paper, we aim to leverage deep learning models to improve users' demography prediction performance. 
To start with, we used multilayer perceptron (\mlp), a class of feed-forward neural net with more than three layers, with embedding as features. In addition, we experimented with
a convolutional neural network (\cnn) ~\cite{kim2014}, which has shown promising and reliable performances across a range of text classification tasks by leveraging embedding results. This is a word embedding based CNN using the text features. 

As a next step we also tried \resnet~\cite{He_2016_CVPR}, whose architecture has more layers (18 for our case), with each layer also connecting to 2 layers in the front. This skip connection adds the ability to train deeper models to avoid overfitting. The accuracy for this model is $1\%$ higher than \mlp with 3 hidden layers. However, \resnet took almost $5$ times time to train. So we decided not to pursue this direction.

\smallskip

\noindent \textbf{Implementation details.} \mlp and \cnn are two advanced models which typically take long times to train over large datasets, especially for the Tumblr data with hundreds of millions of users. To speed up the training process, we sample the same number of female and male examples. Empirical studies over the whole dataset indeed show the model trained from the sampled users can still provide us with convincing performance over the baseline
model (\lr). We implemented \mlp and \cnn using the Keras libary and Tensorflow kernels. For both \mlp and \cnn, we use cross entropy as the objective, and stochastic gradient descent (SGD) as the optimizer. 

\section{Experiments}
\label{section_results}

We conduct
our experiments using the Spark framework on Hadoop with $300$ executors, each of which has  $12G$ memory. \lp is implemented using the Spark GraphX library, while \emb is implemented using a multi-threaded version
of word2vec.


We obtain about $3$ million labeled data as discussed before. To evaluate the performance, we randomly sample $70\%$ of the labeled data as the training set, and the rest as the testing set. We fit our model on the training set, and compute accuracy on the  testing data. For gender prediction we also report AUC as it is a binary classification task.

\subsection{Results}

In short, we demonstrate that \cnn and \mlp can achieve  $0.38$ of the accuracy for the multi-class (7 classes) age prediction, which outperforms the baseline
model by $81\%$ in accuracy. In addition, we show that adding features from \emb and \lp can greatly improve AUC for gender prediction by $5\%$ compared to the baseline
model. Finally, we explore different performance of \lp  as iteration $K$ and hyperparameter $\alpha$ change, as well as the variants of \lp.

\smallskip
\noindent \textbf{Age Prediction.} As discussed above, for the age prediction, we split ages into $7$ buckets $(<17, 18-24, 25-34, 35-44, 45-54, 55-64, 65+)$ as labels, and use \cumfeature and \emb as features. The baseline 
is a generalized multiclass \lr, which uses softmax instead of logistic function as the objective, and \cumfeature.
In addition to \cnn and \mlp, we also combine them together (\cnn+\mlp) to boost the performance, by concatenating the last hidden layers of both and feeding to them to the cross entropy optimizer as the output. Table~\ref{table:age-results} shows the results: combining \cnn and \mlp produces the best results by $81.3\%$ of performance improvement in accuracy compared to the baseline
model \lr. In addition, \mlp outperforms \cnn by $10\%$. 

\begin{table}[t]
	\centering
	\caption{Performance of age prediction.}
	\label{table:age-results}
	{
		\begin{tabular}{ccc}
			\hline
		& Accuracy & Cross-Entropy Loss	\\
			\hline
	\cnn+\mlp	& \textbf{0.3897} & {1.5066}\\ 
			\hline
	\mlp	& 0.3841& 1.5212\\ 
			\hline
    \cnn	& 	0.3482& 1.6157\\ 
			\hline
	\lr & 0.2150 & - \\
	(baseline with \cumfeature) &  &  \\
	\hline
		\end{tabular}
	}
\end{table}


\smallskip
\noindent \textbf{Gender Prediction.} Table~\ref{table:all-results} shows the results of using different features on the Hadoop system. 
\cumfeature is the result that only uses cumulative features~\cite{Grbovic:2015}, which are used in the baseline.
\emb and \lp are the results that use the embedding and label propagation features respectively,  \cumlp is the result that combines both. \all is the result that puts  cumulative, node embedding and label propagation features together.  First, \all produces the best AUC by $5\%$ of performance improvement. Second, it is interesting that \cumlp also gives us competitive results: it achieves only $0.2\%$ of the performance loss compared to \all. Finally, when we separate features respectively, \lp outperforms \cumfeature and \emb by $2\%$.

\begin{table}[t]
	\centering
	\caption{Performance (AUC) of \lr on Hadoop.}
	\label{table:all-results}
	{
		\begin{tabular}{ccccc}
			\hline
				\cumfeature  & \emb & \lp &  \cumlp& \all \\
			\hline
		 0.8489	&0.8413&0.8628&{0.8831}& \textbf{0.8858}\\
			\hline
		\end{tabular}
	}
\end{table}

In addition to testing our model on Hadoop, we also use the following classifiers as baselines for the gender prediction on a single machine. Table~\ref{table:classifiers} shows the quality {(quantifying comparisons)} of different feature integration approaches in terms of AUC and accuracy.
For \emb, we embed users into $50$ dimensional feature space, while for \lp, we are able to obtain $5$ features.  First, our model \nn outperforms other classifiers by up to $10\%$ of performance improvement in both AUC and accuracy. Second, stacking \emb and \lp features together can improve AUC and accuracy (up to $5\%$ of the improvement). Compared to the baseline \lr model with cumulative features, \nn with 
\emb and \lp features improved the AUC by 5\%.


\hide{
Table~\ref{table:all-results} shows the results of using the yamall\footnote{\url{https://github.com/yahoo/yamall}} implementation of \lr on the Hadoop system. \cumfeature is the result that only uses cumulative features~\cite{Grbovic:2015}, \emb and \lp are the results that use the embedding and label propagation features respectively,  \cumlp is the result that combines cumulative  and label propagation features, and \all is the result that puts  cumulative, node embedding and label propagation  features together.  First, \all produces the best AUC by $5\%$ of performance improvement. Second, it is interesting that \cumlp also gives us competitive results: it achieves only $0.2\%$ of the performance loss compared to \all. Finally, when we separate features respectively, \lp outperforms \cumfeature and \emb by $2\%$.

\begin{table}[t]
	\centering
	\caption{Performance (AUC) of \lr on Hadoop.}
	\label{table:all-results}
	{
		\begin{tabular}{ccccc}
			\hline
				\cumfeature  & \emb & \lp &  \cumlp& \all \\
			\hline
		 0.8489	&0.8413&0.8628&{0.8831}& \textbf{0.8858}\\
			\hline
		\end{tabular}
	}
\end{table}

In addition to testing our model on Hadoop via yamall, we also conduct experiments for a variety of classifiers on a single machine. Table~\ref{table:classifiers} shows the quality {(quantifying comparisons)} of different feature integration approaches in terms of AUC and accuracy for state-of-the-art classification models. Note that for \emb, we embed users into $50$ dimensional feature space, while for \lp, we are able to obtain $5$ features. First, \nn outperforms other classifiers by up to $10\%$ of performance improvement in both AUC and accuracy. Second, \lr as a relatively simple model, gives competitive results compared to \xgboost and \gbdt. Finally, stacking \emb and \lp features together can improve AUC and accuracy (up to $5\%$ and $3.7\%$ of the improvement in AUC for \xgboost and \lr respectively). Compared with the baseline \lr model with cumulative features, \nn with 
\emb and \lp features improved the AUC by 5\%.
}

\begin{table*}[t]
\caption{Comparison over various classifiers. Left: AUC; Right: Accuracy.}
\label{table:classifiers}
	\centering

		\begin{tabular}{cccc | cccc}
			\hline
			AUC & \emb & \lp & \emblp & Accuracy & \emb & \lp & \emblp\\
			\hline
		\nn	&  0.883 &	0.865&	\textbf{0.899} & 	\nn	&0.811& 0.804&\textbf{0.830}  \\
			\hline
	\lr	&   0.842 &	0.864 &0.869 & \lr	&0.782&0.783& 0.806\\
			\hline
		\xgboost	&    0.841 &	0.865 &	0.879 & \xgboost	&0.777&0.804&0.812 \\
			\hline
	\gbdt		&  0.831&	0.865 &	0.882  & 	\gbdt		&0.768&0.804& 0.815\\
			\hline
	\linearsvm		&  0.762 &	0.784 & 0.798  & 	\linearsvm		&0.779&0.772& 0.804\\
			\hline
		\end{tabular}
	
\end{table*}

\smallskip

\noindent \textbf{Sensitivity of \lp.} We also evaluate the sensitivity of \lp for hyperparameter $\alpha$ and iteration number $K$. 
Table~\ref{table:varyK} shows the results. 
First, when iteration number $K=1$, \lp just queries labels from 1-hop neighbors, hence, the performance is not as good as examining multi-hop neighbors. Second,  as the number of iterations goes up, the AUC first increases and then slightly decreases. This empirically confirms our intuition that  labels that are close to a node $u$ make more contributions to the labeling result of $u$.
In addition, we notice that the convergence rate is different for different parameters. In general, it takes around $K=50$ iterations for \lp to converge. However, in practice, since we are interested in the prediction performance, we only need to run a few iterations to obtain the best result.  
With regard to the parameter $\alpha$, we observe that the smaller $\alpha$ is, the fewer iterations are needed to achieve the best AUC. For example, when $\alpha = 0.2$, we only need $2$ iterations, while $10$ iterations are needed for the best AUC when $\alpha = 0.8$.

\begin{table*}[t]
	\centering
	\caption{Performance (AUC) of \lp as iteration $K$ and hyperparamter $\alpha$ change.}
	\label{table:varyK}
	{
		\begin{tabular}{c|cccccccccccc}
			\hline
	\diaghead{Alphak}{$\alpha$}{$K$} & 1& 2 &3 &4& 5&6 &7&8&9 & 10 & 15 & 20\\
			\hline
	$ 0.2$	& 0.770&\textbf{0.881} &0.878 &0.878 &0.871 & 0.868& 0.866& 0.865&0.863 &0.862&0.860&0.858 \\
			\hline
				$0.5$	&0.770& 0.853&0.873 &\textbf{0.877}&0.877& 0.876& 0.874& 0.872&0.870&0.868&0.863&0.861 \\
			\hline 
			$0.8$	&0.770&0.816&  0.834& 0.848&0.858& 0.865& 0.869& 0.872&0.874&\textbf{0.875}&0.875& 0.872\\
			\hline
		\end{tabular}
	}
\end{table*}

\hide{
\noindent \textbf{Discussion of \lp.} 
In the demography prediction task, one essential requirement is that \lp can propagate labels all over the graph, as missing values can hurt our performance. Due to the ``small world" phenomenon in social networks, the diameter of the Tumblr graph is small. Hence, we expect \lp to label most nodes
in the small number of iterations. As shown in Table~\ref{table:coverage}, this is indeed the case: even $2$ iterations can label $96.6\%$ of nodes, while we only need $5$ iterations to get $99.1\%$ of nodes labeled. We believe unlabeled nodes are isolated nodes without any following relations.

\begin{table}[t]
	\centering
	\caption{ Coverage of \lp. $R$: the percentage of nodes that are labels.}
	\label{table:coverage}
	{
		\begin{tabular}{c|cccccccc}
			\hline
		{$K$} & 1& 2& 3 &4&5 & 10  & 20\\
			\hline
		$R$ &0.238 &0.966& 0.989& 0.990&0.991&0.991&0.991\\
			\hline
		\end{tabular}
	}
\end{table}
}

\smallskip

\noindent \textbf{Variants of of \lp.}
The variants of \lp proposed in Section~\ref{sec:proposed} are used to examine how the distance of neighbors affects results, and how many iterations are sufficient to get good prediction. Different from the \lp with the constant $\alpha$, both $\beta$-strategy and $\gamma$-strategy decrease the weight on neighboring labels.  As shown in Table~\ref{table:strategy}, both strategies get the best AUC around $4$-$5$ iterations, which suggests that labels within $4$ or $5$-hops are very important for \lp to obtain good results. As the iteration number increases, the AUC almost remains the same especially for the $\gamma$-strategy, which suggests that labels with large distance make little impact on the prediction.

\begin{table}[t]
	\centering
	\caption{Results (AUC) of \lp Variations. $K$ is the number of iterations, $\beta = 0.8$, and $\gamma= 0.9$}
	\label{table:strategy}
	{
		\begin{tabular}{c|ccccccc}
			\hline
			{$K$} & 1& 2& 3 &4&5 & 10\\
			\hline
			$\beta$-strategy &0.770 &0.850& 0.871& 0.877&\textbf{0.878}&0.869 \\
			\hline
			$\gamma$-strategy &0.770&0.868& 0.877& \textbf{0.878}&0.878&0.875 \\
			\hline
		\end{tabular}
	}
\end{table}

\section{Related Work}
\label{sec:related}


\smallskip
\noindent \textbf{Demographic Ad Targeting.}
Personalized advertising is a common ads targeting strategies, which has been studied broadly~\cite{essex2009matchmaker,broder2008computational,majumder2013know}. It tries to display the most relevant ads to each individual. 
Demographic ad targeting is one of the personalization tactics, which effectively targets users based on their age, gender, etc.  Grbovic et al.~\cite{Grbovic:2015} first studied the gender based ad targeting in Tumblr. However, their model is based on learned gender labels, which sometimes can be unreliable. To address their issue, we provide high accuracy labels to improve gender prediction.
In addition to demographic ad targeting, another type of personalized advertising used in Tumblr is interest targeting which recommends ads based on their categories~\cite{Grbovic:2015,shin2014recommending,barbieri2014follow}.

\smallskip
\noindent \textbf{Network Embedding.}
The graph embedding problem tries to generate vector representation of nodes. Previous work such as  locally linear embedding~\cite{roweis2000nonlinear},  IsoMap~\cite{tenenbaum2000global}, and spectral techniques~\cite{chung1997spectral, bach2003learning, belkin2001laplacian}, 
treats networks as matrices. They tend to be slow and do not scale to large networks 
Recently, leveraging  the deep learning techniques, several novel algorithms were proposed to learn feature representations of nodes~\cite{perozzi2014deepwalk, tang2015line, wang2016structural, grover2016node2vec}.
DeepWalk \cite{perozzi2014deepwalk} and Node2Vec \cite{grover2016node2vec} extends the word2vec model~\cite{mikolov2013distributed} to networks by leveraging random walks to generate ``word context". SDNE \cite{wang2016structural} and LINE \cite{tang2015line} learn embedding results on directed graphs that maintain the  first and second order proximity of nodes.  In this paper, our \emb implementation leverages Line to learn graph features of users.


\smallskip
\noindent \textbf{Label Propagation.}
The idea of label propagation has been widely studied in the machine learning literature.  Zhu et al.~\cite{zhu2003semi} first leveraged label propagation for a graph based semi-supervised learning algorithm, and Zhu et al.~\cite{zhou2004learning} further proposed an iterative algorithm from a close form solution for label propagation. After that label propagation has been applied to many domains, such as 
multimedia including image and video data~\cite{lee2013graph} 
and information retrieval like  relevance and keyword search~\cite{he2004manifold, tong2005graph}.  In addition,  Rao and Yarowsky~\cite{rao2009ranking}  proposed a parallel label propagation algorithm under the MapReducde framework. However, none of the above work studied the problem with large-scale data like ours, and implemented the label propagation algorithm on large distributed systems  in practice, nor were these
applied to demography classification.

\section{Conclusion and Future Work}
\label{sec:conclusion}

In this paper, we study the important and challenging problem of Tumblr age and gender prediction for large scale Tumblr data, which can be used to better target sponsored ads against specific demographic audiences. We propose to add the following graph information to the existing cumulative features in Tumblr to enhance the age and gender prediction performance. In particular, we leverage graph embedding and label propagation techniques to generate informative user features, and apply deep learning models including CNN and MLP to utilize these features.
Experimental results demonstrate that our approaches outperforms the 
baseline
models 
by relatively $81\%$ of accuracy improvement for age, and $5\%$ of accuracy and AUC improvement for gender. 



As future work, we would like to incorporate Tumblr's raw age signals either as a feature, or as a way to
gain more trustworthy labels in the age prediction model training.
In addition, the Tumblr data consists of many images. If we could leverage the images and their annotations as extra features, then this could possibly boost the performance further.


\hide{
\begin{itemize}
		\item Develop scalable network embedding approach on Hadoop.
	\item Adapt our approach to age prediction for ads targeting.
	\item Implement \nn on Hadoop, and push it to production.
\end{itemize}
}

\bibliography{ref}
\bibliographystyle{IEEEtran}
\end{document}